# Spontaneous Emotion Recognition from Facial Thermal Images

*Report submitted in fulfillment of the Internship in*

Electrical Engineering Department

by

**CHIRAG KUMAR KYAL**

**ID: 20199907341273428**

**Email: chirag.kyal@gmail.com**

*under the guidance of*

**Prof. Aurobinda Routray**

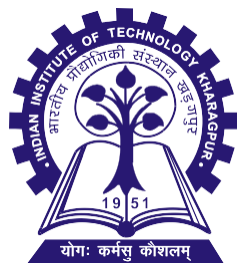

**Electrical Engineering Department**
Indian Institute of Technology, Kharagpur, India July 2019

**Assignment:1**

# Deep Learning and Cuda-Computing method for Facial Landmarks and Emotion Labels prediction on Thermal Images

*Introduction*:
One of the key research areas in computer vision addressed by a vast number of publications is the processing and understanding of images containing human faces. The most often addressed tasks include face detection, facial landmark localization, face recognition and facial expression analysis. Other, more specialized tasks such as affective computing, the extraction of vital signs from videos or analysis of social interaction usually require one or several of the aforementioned tasks that have to be performed. In our work, we analyze that a large number of tasks for facial image processing in thermal infrared images that are currently solved using specialized rule-based methods or not solved at all can be addressed with modern learning-based approaches. We have used USTC-NVIE database for training of a number of machine learning algorithms for facial landmark localization.

*Methodology:*
All selected frames of USTC-NVIE database were manually annotated with the 68-point landmark set. This extensive set of annotations using a widely established scheme allows using the database for a substantial number of algorithms, allowing assessment of their performance on thermal infrared data. Figure 1 shows examples of annotated frames while Figure 2 shows the exact localization of the 68 landmark positions in the face. Both the landmarks as well as the connectivity information are stored, allowing selection of landmarks of specific facial areas such as eyes or mouth separately. After landmarking all images, the dataset has been checked for annotation consistency, ensuring that landmark positions correspond to the same facial features in all database images.

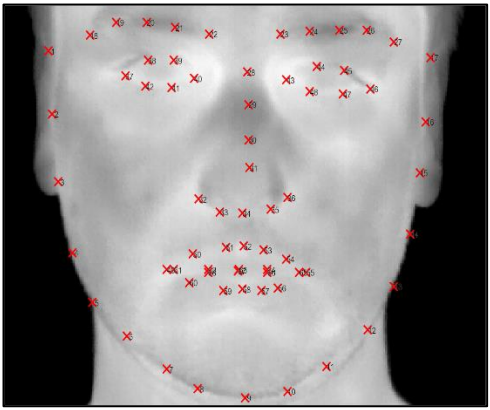
*Fig:1* The 68-point annotation scheme with each point's coordinates in the face.

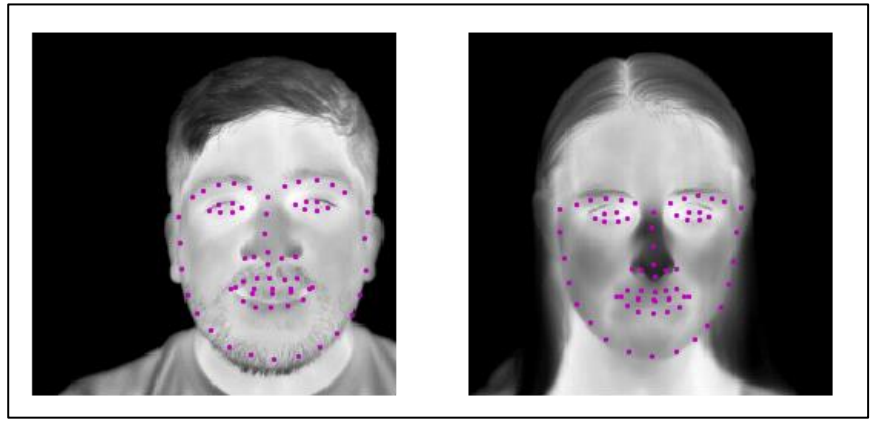
*Fig:2* Sample images from the database with the 68-point landmarks shown as overlay.

Deep learning - the use of multilayer neural networks for machine learning tasks - is the currently dominating research area in image processing. The most commonly used networks in computer vision are various types of convolutional neural networks, a sub-type of neural networks that uses locally connected convolutional layers in addition to (or as replacement for) classical fully connected layers. Convolutional networks have achieved outstanding performance in tasks such as image classification, where residual networks or the recently proposed Mask R-CNN are the current benchmark for multi-label classification. Fully convolutional networks such as the U-Net are the current state-of-the-art in image segmentation. Other popular ImageNet models such as VGG16, Inception Net, Res Net are also used for training purpose but all of them unable to give proper fitting accuracy. Deep learning techniques have also been applied to the task of facial landmark detection, where a deep network is trained to detect a set of facial landmarks together with additional attributes such as head pose, gender and basic facial expressions.

To analyze the suitability of the database for deep learning tasks, we have chosen to train and evaluate the recently proposed *deep alignment network (DAN)* algorithm (Fig:3) with the database. The deep alignment network is a multistage approach in which several stages refine landmark positions predicted by the previous stage. The initial stage receives an input image and as output it yields a transformed version of the image warped into a normalized canonical form together with a first estimate of the facial landmarks as a heatmap and 256 features from the last, fully connected layer of the first stage. After the image has been transformed by the first stage, all subsequent stages have an identical layout. Consequently, their input and output are a canonical image, a landmark heatmap and the feature map. We implemented only one single second stage after the initial stage as further stages drastically increase the required training time while are reported to yield no improvement in detection precision. The network returns the landmark positions and not a full-face model. However, since the goal of the algorithm is the landmark detection and improving fitting accuracy.

CUDA- is a parallel computing (Fig:4) platform and application programming interface (API) model created by Nvidia. It allows software developers and software engineers to use a CUDA-enabled graphics processing unit (GPU) for general purpose processing — an approach termed GPGPU (General-Purpose computing on Graphics Processing Units). The CUDA platform is a software layer that gives direct access to the GPU's virtual instruction set and parallel computational elements, for the execution of compute kernels. The above Deep Alignment Network is accelerated by using CUDA Computing over PyTorch so that the training, validation and testing part can be completed in less time. In GPU-accelerated applications, the sequential part of the workload runs on the CPU – which is optimized for single-threaded performance – while the compute intensive portion of the application runs on thousands of GPU cores in parallel.

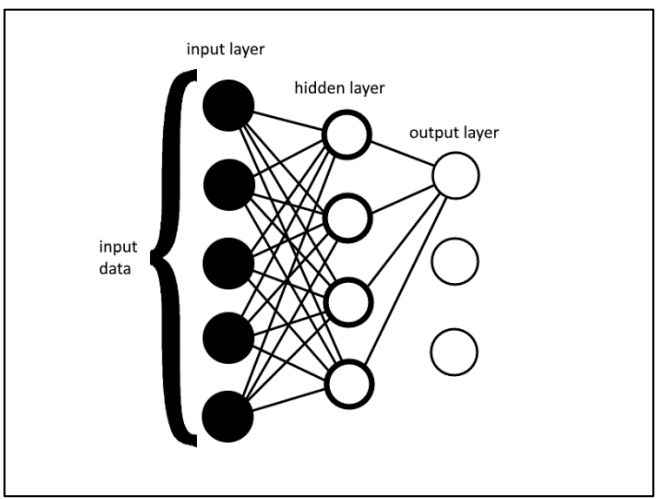

*Fig:3. Deep alignment network (DAN) model architecture*

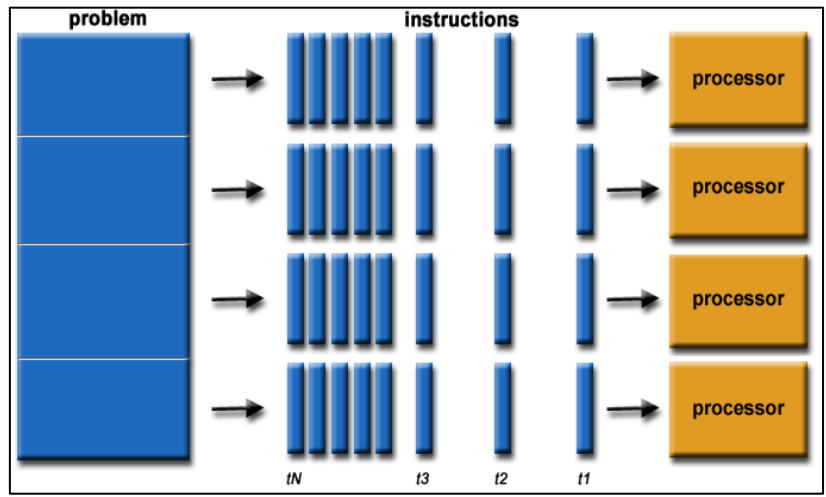

*Fig:4. Parallel Computing and CUDA Processing Workflow*

*Experimental Result:*
Here, we describe how the algorithms for each task were trained and evaluated. Results of the facial expression analysis evaluation was obtained using leave one-subject-out-cross validation. In this validation type, we removed all images of a given subject from the database, trained the algorithms on all remaining subjects and tested their performance on the subject previously removed from the database. While requiring a large number of evaluation runs, this method was chosen as it gives the best impression of the overall algorithm and database performance due to the maximal possible overlap of training data with the full database while still allowing evaluation on unseen subjects. We did not perform the same type of cross-validation for the facial landmark detection as the DAN algorithm requires substantially longer for training its neural network than the other used algorithms; training a DAN instance requires about 72 hours on a GeForce 980Ti GPU. Therefore, we randomly picked 270 images from 8 subjects from the database as test set and trained the landmark detection algorithms on the remaining images to perform evaluation within reasonable time.

Figure:5 deploys the comparison between different deep learning models used to test the result of Face Landmarking on the database. It can be easily visualized from that Inception Net (fig:5 (b)) has given the least good result as compared to other models with an accuracy of less than 40%, whereas VGG16 and ResNet have given better result with an accuracy of about 60% and 65% respectively. However, the prediction accuracy of the Face landmark annotation in our proposed model (DAN) is above 80% for all the tested images. Therefore, in our experiment DAN has given better result for this problem.

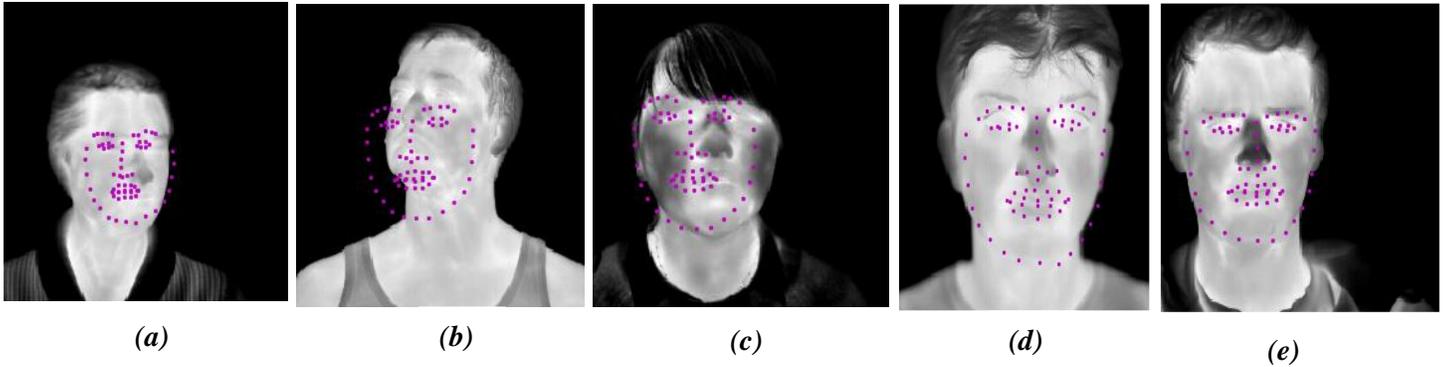

*Fig:5 Testing Output (a)VGG16, (b) Inception Net, (c) ResNet, (d) & (e) Deep alignment network (DAN)*

*Conclusion:*

In our work, we have manually annotated the high-resolution thermal face image database for different computer vision tasks and evaluated how different algorithms perform on commonly appearing problems when trained using the database. We have thoroughly described the database's image acquisition procedure and its contents. Afterwards, we can evaluate its suitability for facial landmark detection and facial expression recognition. We

| Deep Learning Models | Time Required for training and Testing | Accuracy % |
|---|---|---|
| DAN | 72 hrs | > 80% |
| VGG16 | 78 hrs | ~ 60% |
| ResNet | 76 hrs | ~ 65% |
| Inception Net | 65 hrs | < 40% |

were able to show that both tasks can be solved robustly by using learning-based approaches that are trained using the database. Evaluation has shown that the learning-based approaches, several of which have not been used for these tasks in the thermal infrared domain before, clearly outperform previously presented methods. In conclusion we were able to show that using a sufficiently large and well annotated database can be used to train different learning-based algorithms which should be preferred over algorithm-based approaches due to their increased performance.

# Assignment:2
# Feature-based Face tracking and Face Alignment
# for Spontaneous Thermal Images

*Introduction*:

Feature tracking is one of fundamental steps in many computer vision algorithms and the KLT (Kanade-LucasTomasi) method has been successfully used for optical flow estimation. Feature tracking is the foundation of several high-level computer vision tasks such as motion estimation, structure from motion, and image registration. Since the early works of Lucas and Kanade and Shi and Tomasi, the Kanade-Lucas-Tomasi (KLT) feature tracker has been used as a de facto standard in handling point features in a sequence of images. From the original formulation a wide variety of extensions has been proposed for better performance. Baker and Matthews summarized and analyzed the KLT variants in a unifying framework. KLT makes use of spatial intensity information to direct the search for the position that yields the best match. It is faster than traditional techniques for examining far fewer potential matches between the images.

*Methodology:*

The KLT is a local minimizer of the error function e between a template T and a new image I at frame t + 1 given the spatial window W and the parameter p at frame t.

$$e = \sum x \in W \; [T(x) - I_{t+1}(w(x; p_t, \delta p))]^2 \quad (1)$$

Conventionally the Gauss-Newton method is used to search the parameter change δp in this minimization. The KLT variants differ by the tracking motion model w it employs and the way to update the parameter from δpt. For example, the translation model is defined with a two-dimensional translation vector p.

$$w(x, p) = x + p \quad (2)$$

We can expect computational efficiency by switching the roles of the image I and the template T in (1).

$$e = \sum x \in A \; [I(w(x, p_t)) - T(w(x; \delta p))]^2 \quad (3)$$

The first-order Taylor expansion of (3) gives

$$e \approx \sum x \in A \; [I(w(x, p_t)) - T(w(x; 0)) - J(x)\delta p]^2 \quad (4)$$

where the Jacobian $J = \frac{\partial T}{\partial p}|p = 0$. With the Hessian approximated as $H = \sum J^T J$, the local minimum can be found by minimizing (4) iteratively

$$\delta p = H^{-1} \sum x \in A \; J^T [I(w(x, p_t)) - T(x)] \quad (5)$$

with the parameter update rule

$$w(x; p_{t+1}) = w(x; pt) \cdot w(x; \delta p)^{-1} \quad (6)$$

This is called inverse compositional image alignment and it takes advantage of a single Hessian computation only when a feature is registered.

*Objective:*

The main objective to use KLT Tracker algorithm is to make all the frontal faces of USTC-NVIE database in a same alignment (Fig:1) so that the DAN (previous assignment) algorithm can be applied on all the spontaneous faces and face landmarks can be detected automatically. Figure 3 deploys the principle of frame by frame feature points tracking.

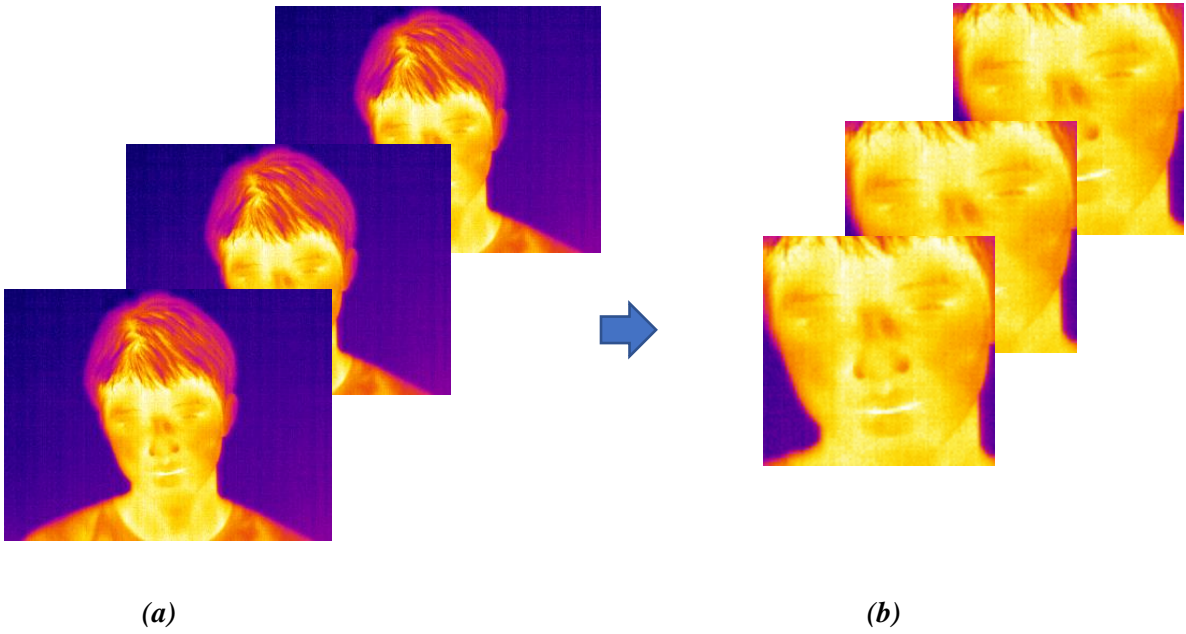

*(a)*                      *(b)*

***Fig:1*** *(a) Spontaneous USTC-NVIE data images. (b) USTC-NVIE images with aligned faces*

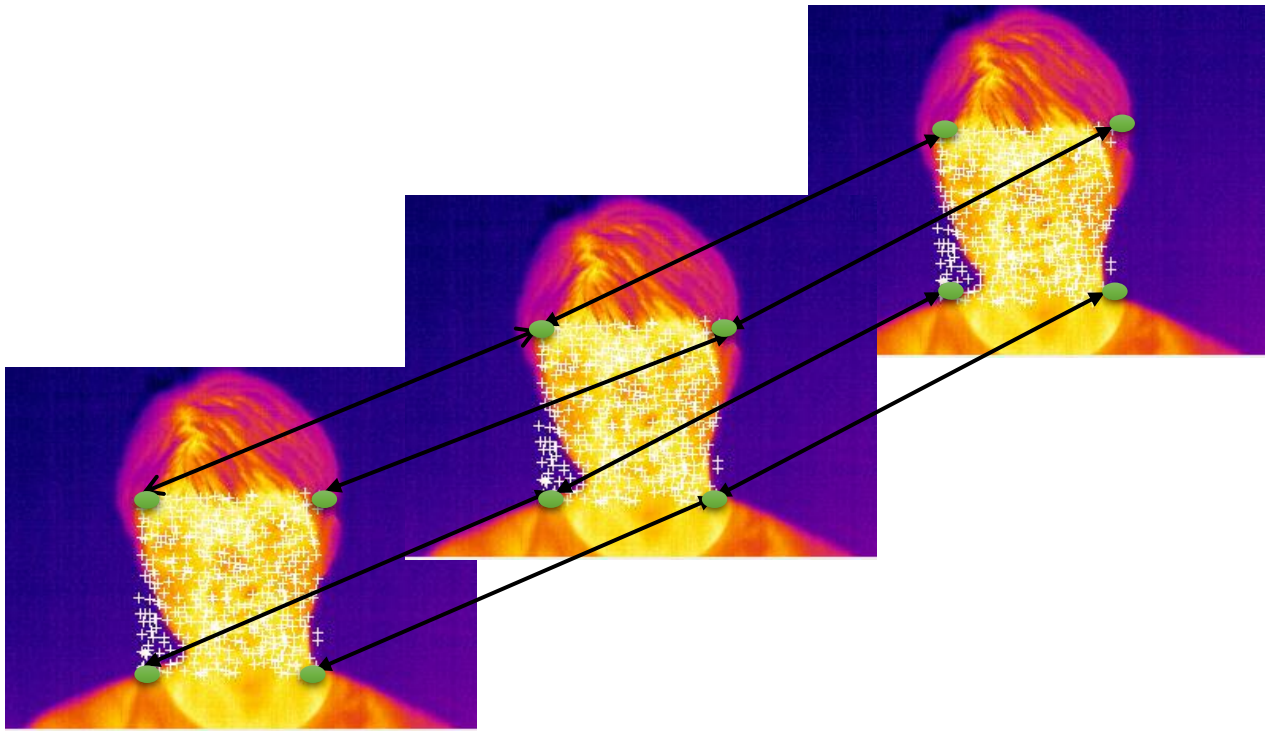

***Fig:2*** *Frame by Frame feature Points Tracking*

*Experimental Result:*
The KLT tracker works in three steps:
1. Detection of Human face (Fig 3)
2. Feature points detection on the face (Fig 4)
3. Tracking of feature points over the spontaneous thermal images (Fig 5)

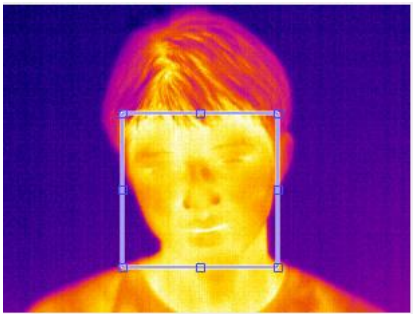
*Fig:3* Face Detected

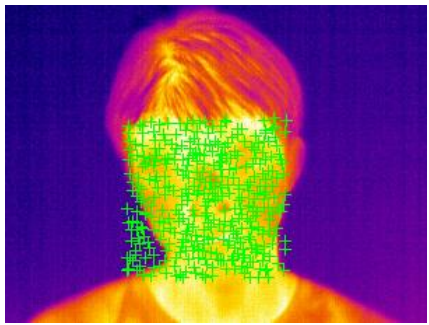
*Fig:4* Feature points detected

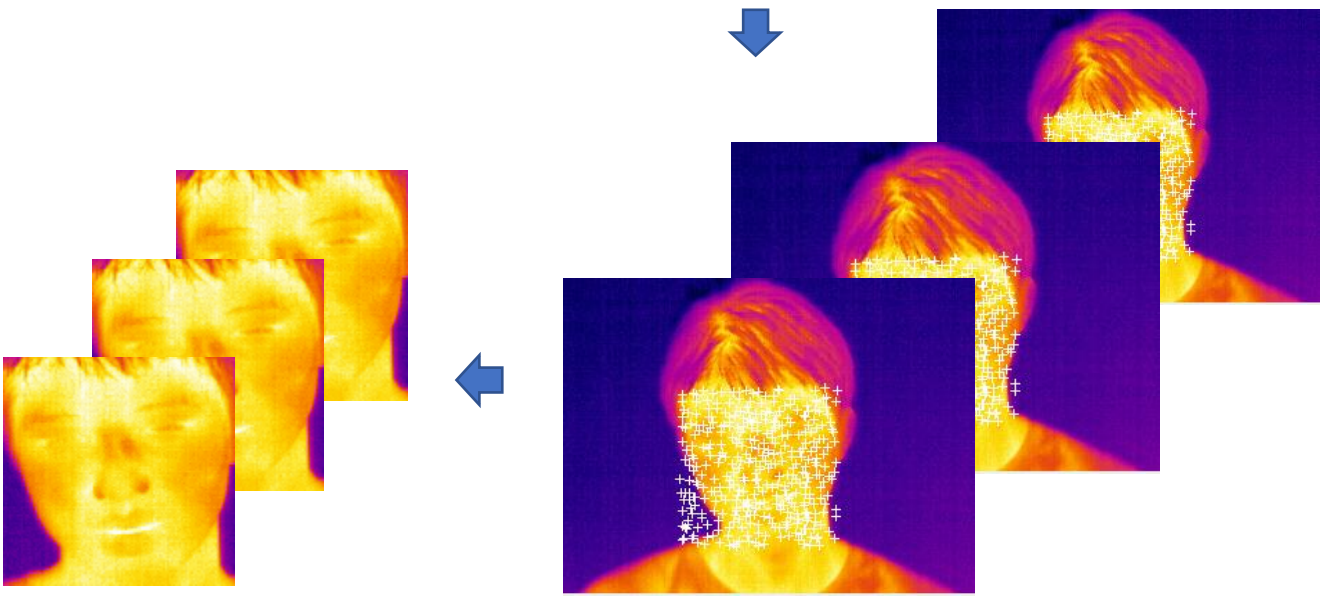

*Fig:6* Final Aligned Faces

*Fig:5* Feature Points Tracking

*Conclusion:*
We have implemented a feature tracking and optical flow detecting algorithm – KLT Tracker. This tracker is used to make all the frames of the USTC-NVIE database in a same alignment so that we can easily detect the feature points on it using DAN algorithm. Figure 6 represents the Flow Chart of the KLT Tracker algorithm.

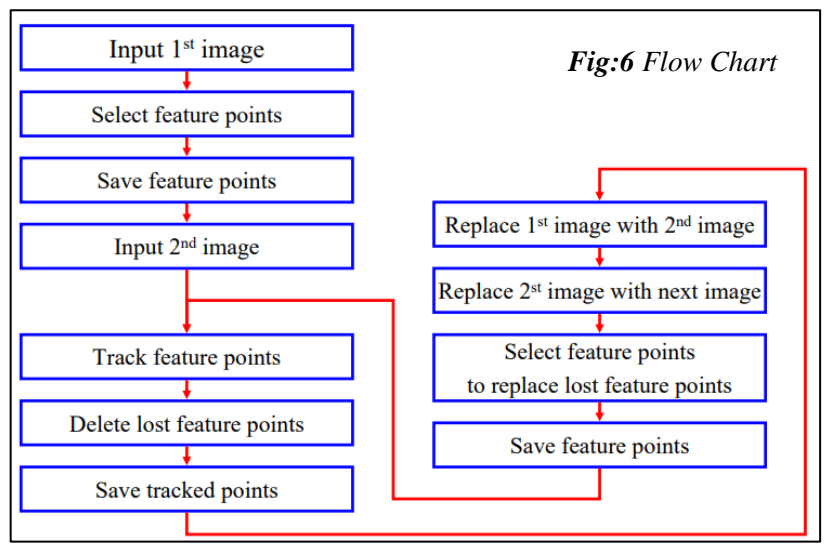
*Fig:6* Flow Chart

# Assignment:3
## Thermal Features Extraction from ROI Patches

*Introduction*:
There are numerous works have been done in Facial Emotion Recognition (FER) in visible domain. However, visible imagery is incapable of discerning the true human emotions. In other words, people can conceal their true emotions due to circumstantial reasons. Hence the other possible means is to study the thermal signature of the facial regions as a spontaneous feature to detect true emotion. Unlike visible imagery, the pixel intensity of thermal images corresponds to surface temperature and emissivity of the object. The intensity value is directly proportional to the surface temperature of the object (in our case the object is face). The change in the surface temperature of the skin is mainly caused due to blood flow through superficial blood vessels. Limited attempts have been done in the area of Thermal Imagery based Emotion Recognition (TIER).

*Motivation*:
We obtained the sensitive portions of the face which undergoes prominent changes (increase or decrease of thermal intensity) during the emotional elicitation using the ratio-based method. The chosen regions are forehead, nose, maxillary, left cheek and right cheek. For the sake of normalization, we have resized the cropped images into a consistent size. The choice of size for the respective regions are mentioned below. The BSIF filters are trained accordingly with the cropped images for feature extraction purpose. We have used DAN algorithm (discussed in Assignment 1) to auto detect 68 landmark points over the face. After detecting those feature points, we can extract different ROI regions by using simple distance techniques.

*Objective*:
The main objective of this assignment is to extract different region of interests such as forehead, nose, left cheek, right cheek, nose and maxillary. These regions will act as features for emotion detection because the temperature if these regions changes with the temperature change in the face. The temperature of human face is directly impacted by our mental expression and emotions. Human may hide their emotions externally but they can't hide their inner facial emotions. Visual Imagery fails to solve this trap, where the Thermal Imaging comes into picture. Different facial regions change their temperature according to the emotion of the human, different emotions have different temperature variation on the face. Those changes in temperature can be easily detected by using a thermal IR camera. So those ROI need to be extracted properly so that different machine learning techniques can be applied on it to classify those emotions.

*Different ROI regions:*

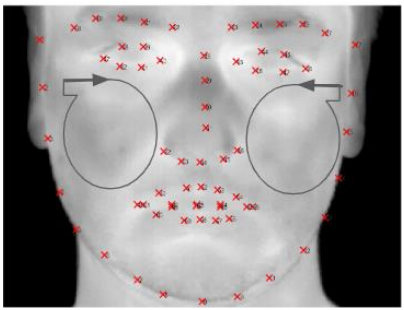

**Region 1: Left cheek:**

2 -> 37 -> 42 -> 41 -> 40 -> 32 -> 50 -> 49 -> 6 -> 5 -> 4 -> 3 -> 2

**Region 2: Right cheek:**
16 -> 46 -> 47 -> 48 -> 43 -> 36 ->64 -> 65 -> 12 -> 13 -> 14 -> 15 -> 16

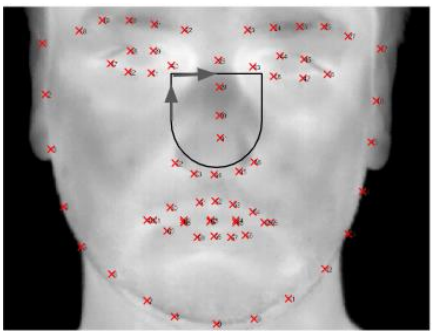

**Region 3 - Nose:**
40 -> 28 -> 43 -> 36 -> 35 -> 34 -> 33 -> 32 -> 40

*ROI regions for different emotions:*

- Happy: left cheek, nose and right cheek

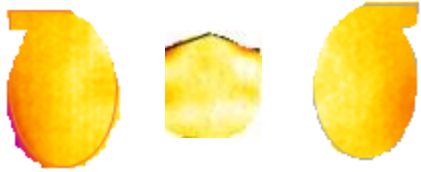 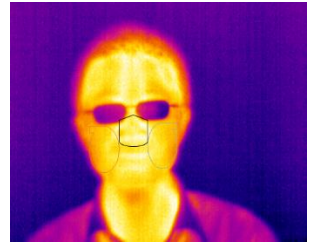

- Disgust: left cheek, nose and right cheek

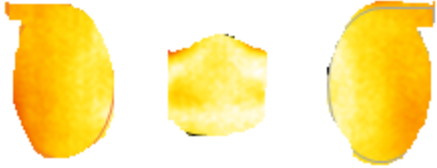 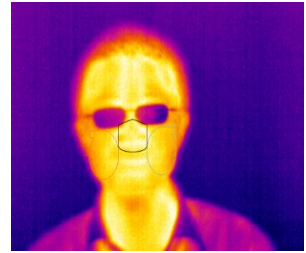

- Fear: left cheek, nose and right cheek

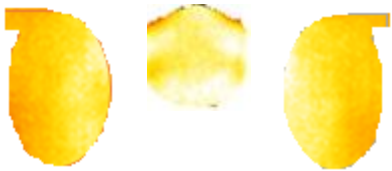 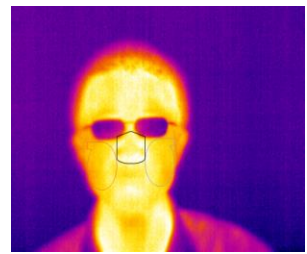

- Surprise: left cheek, nose and right cheek

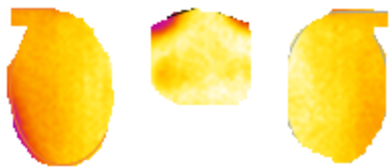 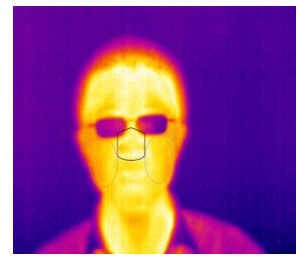

*Conclusion*:
Visible imagery is incapable of detecting the true human emotions. People can conceal their true emotions due to circumstantial reasons. Hence the other possible means is to study the thermal signature of the facial regions as a spontaneous feature to detect true emotion. Unlike visible imagery, the pixel intensity of thermal images corresponds to surface temperature and emissivity of the object. The intensity value is directly proportional to the surface temperature of the object (in our case the object is face). The change in the surface temperature of the skin is mainly caused due to blood flow through superficial blood vessels. Hence, by using different deep learning models we can classify these regions to detect the change in temperature as well as change in emotion of human beings.

# Assignment:4

## Human Emotion Recognition by Combination of Pixel Selection with Covariance Similarity Approach

*Introduction*:

Rich spectral information of thermal images provides a non-invasive way to characterize the skin tissues and thereby improves thermal face recognition accuracy. However, the increased computational complexity is reduced by efficient feature selection method. We amalgamate pixel selection with spectral discrimination. The pixel selection process choses the informative pixels, which improves the computational performance, whereas, covariance similarity encompasses the complete spectral information. We compare the covariance matrices formed from the selected pixels obtained by DAN algorithm (discussed in Assignment 1). A detailed study of the covariance similarity measures has been conducted. The rich spectral information provided by thermal data enhances the discrimination ability, which in turn decreases the inter-personal similarities. We have used covariance measure-based facial emotion recognition strategy, where we compare the utility of common covariance measures. we identify only the important facial points and carry out face emotion recognition task on the reflectance patterns obtained from these points. We exploit the ability of spectroscopy to characterize skin components. Since, the covariance matrix of raw or windowed fiducial point spectra leads to a compact representation of the entire face, the pixel selection process achieves satisfactory recognition result albeit at reduced computational cost.

*Methodology*:

We present a novel covariance similarity approach of thermal facial emotion recognition and include a complete study of the prevalent covariance similarity measures. We identify the 68 fiducial or landmark points by DAN algorithm and calculate the covariance matrix from the raw fiducial point spectra (raw FPS) (Fig.1). The emotion matching procedure involves calculation of similarity between covariance matrix of the probe and the gallery. A schematic of the workflow is shown in Fig. 2.

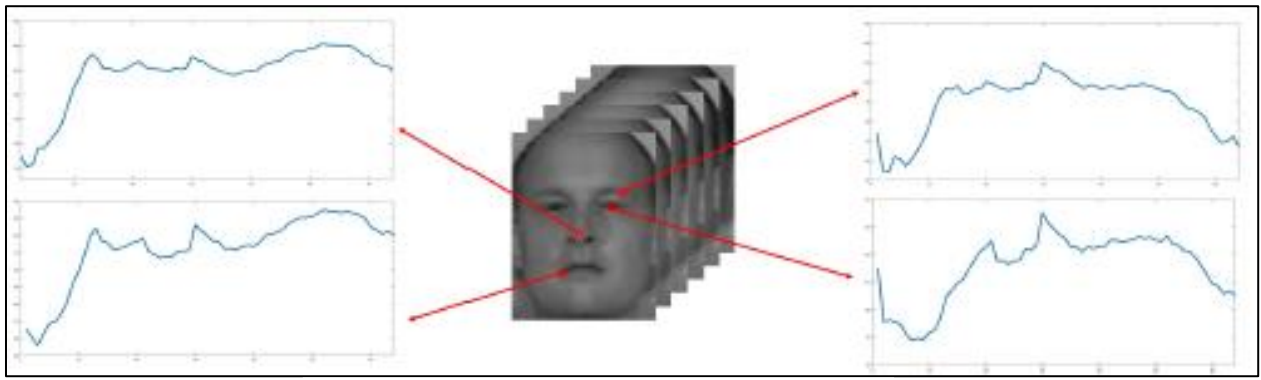

*Fig:1* Raw Fiducial Point Spectra (raw FPS)

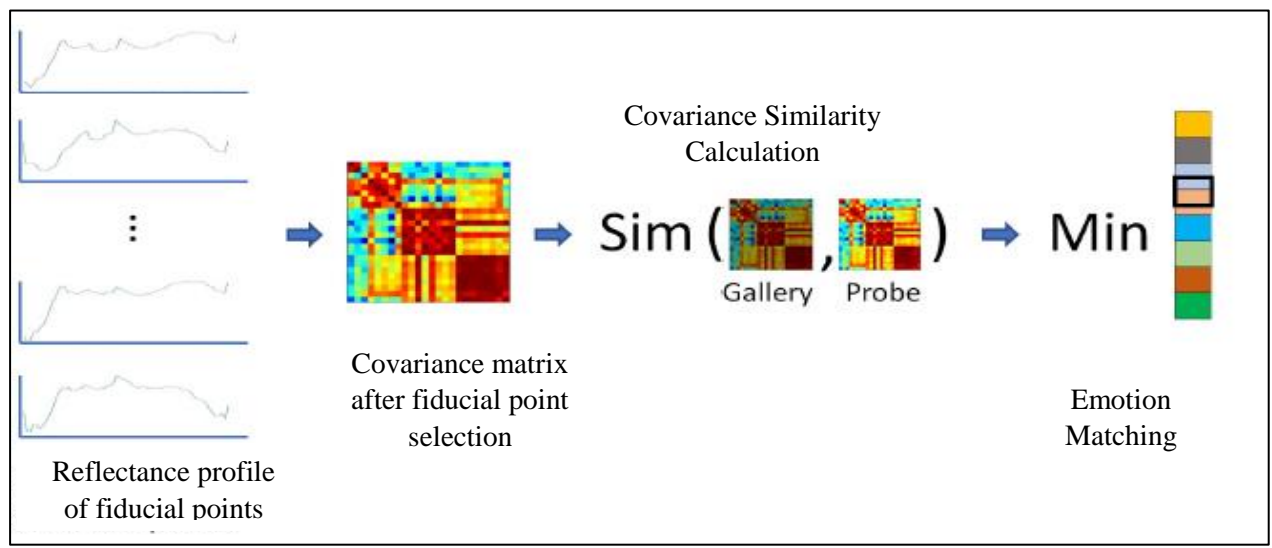

*Fig:2* Proposed Schematic Workflow

> **Covariance Similarity Measure**

Covariance matrix represents the data in a cohesive and compact structure. Covariance matrix is a specific symmetric positive definite matrix, which arises at several application domains such as- human identification, image segmentation, multi-camera tracking etc. We represent each thermal face cube of a specific subject by a covariance matrix, which encodes the spatial as well as spectral details of the data. However, choosing proper measures to compare the similarity between two covariance matrices is a non-trivial task because the covariance spaces form a Riemannian manifold rather than traditional Euclidean space. The similarity measure adopted should ideally consider the manifold geometry into consideration. Assume that we need to compute the similarity between covariance matrices X and Y. The following covariance similarity measures can be employed-

- **Cholesky Distance (CHOL)**

This measure is based on Cholesky distance which assumes that every PSD matrix can be written in terms of inner product of lower triangular matrices, such that- $X = L_1 L^T_1$ and $Y = L_2 L^T_2$. The Cholesky distance between covariance matrices X and Y can be written as-

$$D_{CHOL}(X, Y) = \| L_1 - L_2 \|_F$$

- **Affine Invariant Riemannian Metric (AIRM)**

Affine Invariant Riemannian Metric (AIRM) between two SPD matrices X and Y are calculated by-

$$D_{AIRM}(X, Y) = \| \log(X^{-1/2} Y X^{-1/2}) \|_F$$

The logarithmic term log (.) is the principal matrix logarithm. The disadvantages of using this metric are calculation of the similarity measure is very time consuming as it requires computation of eigenvalues and principal logarithm.

- **Log Euclidean Riemannian Metric (LERM)**

Log Euclidean Riemannian Metric (LERM) between X and Y is calculated by-

$$D_{LERM}(X, Y) = \| \log(X) - \log(Y) \|_F$$

- **Jeffrey's KL Divergence**

Jeffrey's KL Divergence between the covariance matrices X and Y is written as-

$$D^2_{JKLD}(X, Y) = \frac{1}{2} \text{tr}(X^{-1}Y + Y^{-1}X - 2I)$$

where, I is the identity matrix of size L

- **Jensen Bregman Log-Det Divergence (JBLD)**

Jensen-Bregman log-det divergence between two PSD matrices is represented as-

$$D_{JBLD}(X, Y) = \log \left| \frac{X+Y}{2} \right| - \frac{1}{2} \log |XY|$$

where, |.| represent determinant of the matrix.

Each similarity measures have their advantages and drawbacks, and their suitability solely depends on the background and problem sets.

*Experiment and Results*:

A. **Dataset Details**

We tested the proposed methodology on an established dataset named Natural Visible and Infrared facial Expression Database (NVIE) dataset. The dataset provides images of apex and onset emotion of a subject for separate cases. We studied only the apex thermal images for all six emotions viz. happiness, disgust, fear, surprise, anger, and sadness.

B. **Importance of Fiducial point selection**

This section analyzes the importance of fiducial points spectrum (FPS) selection by comparing the covariance similarity. With increase in window size results in higher similarity and affirms the significance of raw or windowed fiducial point selection. The result also confirms the efficacy of JKLD for this particular application. It can be clearly said that increase in window size boosts AIE value. Use of 7_7 window leads to adequate AIE.

### C. Runtime performance evaluation

We compare the runtime performance in Anaconda Spyder 3.2 version on Intel i5 processor having 8GB RAM. The average runtime of 10 iterations is considered for comparison. The lower computational requirement of the proposed method can be mainly attributed to the pixel selection process.

### D. Importance of Confusion Matrix

For each of the Covariance matrix similarity method, a confusion matrix is created for each emotion with all the subjects. For a given unknown emotion, we need to test the distance with all other emotions with the help of confusion matrix. The emotion having least distance with the asked emotion will be chosen as predicted emotion. Fig 3 represents an example of such confusion matrix.

|          | Happy | Disgust | Fear | Surprise | Anger | Sadness |
|----------|-------|---------|------|----------|-------|---------|
| Happy    | 0     | 55      | 63   | 10       | 11    | 36      |
| Disgust  | 55    | 0       | 44   | 5        | 8     | 84      |
| Fear     | 63    | 44      | 0    | 2        | 74    | 41      |
| Surprise | 10    | 5       | 2    | 0        | 85    | 52      |
| Anger    | 11    | 8       | 74   | 85       | 0     | 31      |
| Sadness  | 36    | 84      | 41   | 52       | 31    | 0       |

*Fig:3 An example of Confusion Matrix*

*Conclusion*:

We have used a covariance similarity approach towards thermal facial emotion recognition. Investigation of prevalent covariance similarity measures highlight the usefulness of Jeffrey's KL divergence measure over others. The work demonstrates that the use of fiducial points instead of the whole data is sufficient. The real supremacy of the work lies in the considerable improvement in matching unknown emotion face due to the inherent ability of spectroscopy to characterize human face constituents. The use of fiducial points helps in improved performance in pose invariant matching.

# Assignment:5
## Discriminant neighborhood embedding for classification

*Introduction*:
Manifold learning plays an important role in many applications such as pattern representation and classification. Several nonlinear techniques have been introduced to learn the intrinsic manifold embedded in the ambient space of high dimensionality. However, these nonlinear techniques yield maps that are defined only on the training data points and it remains unclear how to evaluate the maps on novel testing points. To address this problem, locality preserving projection (LPP) was proposed. Since LPP does not make use of class label information, it cannot perform well in classification. Local discriminant embedding (LDE) incorporates the class information into the construction of embedding and derives the embedding for nearest-neighbor classification in a low-dimensional space. Furthermore, both LPP and LDE suffer from singularity of matrix when the number of training samples is much smaller than the dimensionality of each sample. To address this problem, a novel method called discriminant neighborhood embedding (DNE) is proposed. DNE is inspired by an intuition of dynamics theory. Multi-class data points in high-dimensional space are supposed to be pulled or pushed by discriminant neighbors to form an optimum embedding of low dimensionality for classification.

*Discriminant neighborhood embedding:*
Suppose $N$ multi-class data $\{x_1, x_2, \ldots, x_N\}$ are sampled from underlying manifold $M$ embedded in high-dimensional ambient space $\mathbf{R}n$ and any subset of data points in the same class is assumed to lie in a submanifold of $M$. We seek an embedding characterized by intra-class compactness and inter-class separability. The larger the distance between two points is, the weaker the interaction becomes. Mutual force can be distinguished as intra-class attraction or inter-class repulsion between the pair of data points from the same or different class, respectively. Furthermore, one neighbor for a point is referred to as intra-class neighbor if they belong to the same class; and inter-class neighbor otherwise.

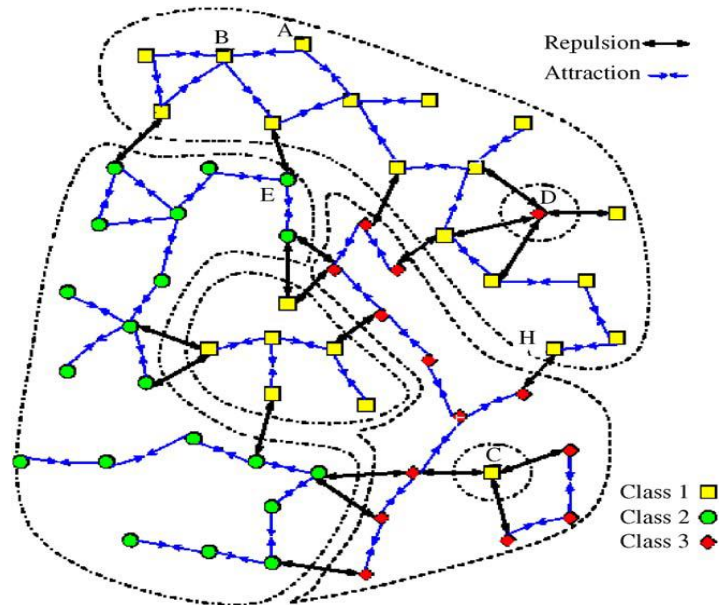

*Fig. 1. Points are pulled and pushed by their neighbors due to the local force of attraction and repulsion, respectively.*

One point whose $k$ nearest-neighbors are all its intra-class neighbors is defined as *inner point*, e.g. *Point A* and *B* in Fig. 1. An inner point is enveloped by the points of the same class and just acted on by local force of intra-class attraction. One point whose $k$ nearest-neighbors are all its interclass neighbors is called *insular point*, e.g. *Point C* and *D* in Fig. 1. An insular point is isolated by the points of different classes and just acted on by local force of inter-class repulsion. A third type of point is called *marginal point*, e.g. *Point E* and *H* in Fig. 1, whose $k$ nearest-neighbors include intra and inter-class neighbors simultaneously. In other words, marginal points are acted on by local intra-class attraction and inter-class repulsion at the same time. The inner point *B* is pulled nearer toward its intra-class neighbors due to attractions whereas insular point *C* is pushed farther away from its inter-class neighbors due to repulsions. As to marginal point, *E* is not only pulled but pushed as well, and moves along the direction of the net force. As a result, points in the same class form compact submanifold while the gaps between submanifolds corresponding to different classes become wider than before.

An edge is put between nodes *i* and *j* if $x_i$ is among $k$ nearest-neighbors of $x_j$ or $x_j$ is among $k$ nearest-neighbors of $x_i$. To distinguish the local intraclass attraction and inter-class repulsion between neighboring points, each edge is weighed +1 or

−1, respectively. Suppose that $c_i$ denotes the class label of $x_i$ and $knn(i)$ denotes the set of $k$ nearest-neighbors of $x_i$, then the adjacent matrix $F$ of graph $G$ which models the underlying supervised manifold structure is as follows:

$$F_{ij} = \begin{cases} +1 & (x_i \in knn(j) \lor x_j \in knn(i)) \land (c_i = c_j), \\ -1 & (x_i \in knn(j) \lor x_j \in knn(i)) \land (c_i \neq c_j), \\ 0 & otherwise \end{cases} \quad (1)$$

We estimate the motion of all points by learning a linear transformation of the input space such that in the transformed space, intra class compactness and inter-class separability are achieved simultaneously. We denote the transformation by a matrix $P$, define intra-class compactness as

$$\Delta(P) = \sum_{i,j}^{\cdot} \left\| P^T x_i - P^T x_j \right\|^2 \quad (x_i \in knn(j) \lor x_j \in knn(i)) \land (c_i = c_j) \quad (2)$$

and similarly define inter-class separability as

$$\delta(P) = \sum_{i,j}^{\cdot} \left\| P^T x_i - P^T x_j \right\|^2 \quad (x_i \in knn(j) \lor x_j \in knn(i)) \land (c_i \neq c_j) \quad (3)$$

then one reasonable criterion for "good" motion is to minimize
$$\varphi(P) = \Delta(P) - \delta(P) \quad (4)$$

It is an indefinite quadratic form. Minimizing criterion (4) is an attempt to minimize the expected distance among the samples of the same class and to maximize the total distance between different classes simultaneously. Using (1), the criterion (4) can also be expressed as

$$\varphi(P) = \sum_{i,j}^{\cdot} \left\| P^T x_i - P^T x_j \right\|^2 F_{ij} \quad (5)$$

Following some simple algebraic steps, we get that

$$\varphi(P) = \sum_{i,j}^{\cdot} \left\| P^T x_i - P^T x_j \right\|^2 F_{ij}$$

$$= 2 \sum_{i,j}^{\cdot} (x_i^T P P^T x_i - x_i^T P P^T x_j) F_{ij}$$

$$= 2 \sum_{i,j}^{\cdot} tr((P^T x_i x_i^T P - P^T x_j x_i^T P) F_{ij})$$

$$= 2 tr \left( \sum_{i,j}^{\cdot} (P^T x_i F_{ij} x_i^T P - P^T x_j F_{ij} x_i^T P) \right)$$

$$= 2 tr(P^T X S X^T P - P^T X F X^T P)$$

$$= 2 tr(P^T X (S - F) X^T P) \quad (6)$$

where $tr(\cdot)$ is the trace of matrix and $X = [x_1, x_2, \ldots, x_N]$. S is a diagonal matrix whose entries are column (or row, since $F$ is symmetric) sums of $F$, i.e $S_{ii} = \Sigma_j F_{ji}$. Then the optimal transformation matrix $P$ can be obtained by

$$P_{optimal} = \operatorname{argmin}_P tr(P^T X(S - F) X^T P) \quad (7)$$

By constraining that $P^T P$ is an identity matrix, we switch to a simple eigenvalue and eigenvector problem with respect to the symmetric matrix $X(S−F)X^T$. Because the quadratic form $\varphi(P)$ is indefinite, the eigenvalues of $X(S − F)X^T$ may be positive, negative or zero. Suppose d is the number of the negative eigenvalues, $\lambda_1 \leq \cdots \leq \lambda_d < 0 \leq \lambda_{d+1} \leq \cdots$, then the transformation matrix P is constituted by the d eigenvectors of $X(S − F)X^T$ corresponding to its first d negative eigenvalues. Once the n × d transformation matrix P has been learnt, the embedding of any new test sample $x_{new} \in R^n$ is accomplished by $y_{new} = P^T x_{new}$, where $y_{new} \in R^d (d<<n)$.

## Local Phase Quantization (LPQ):

The Local Phase Quantization descriptor, has been used to find out a feature vector of length 256 from an input image. This feature vector will act as a node in the Graph G. LPQ is constructed to retain an image in the local invariant information to artifacts generated by different forms of blur. Inspired by this idea, we have used the LPQ as an effective method to solve the problem of expressions variations. The following chart shows the steps necessary to build the LPQ descriptor.

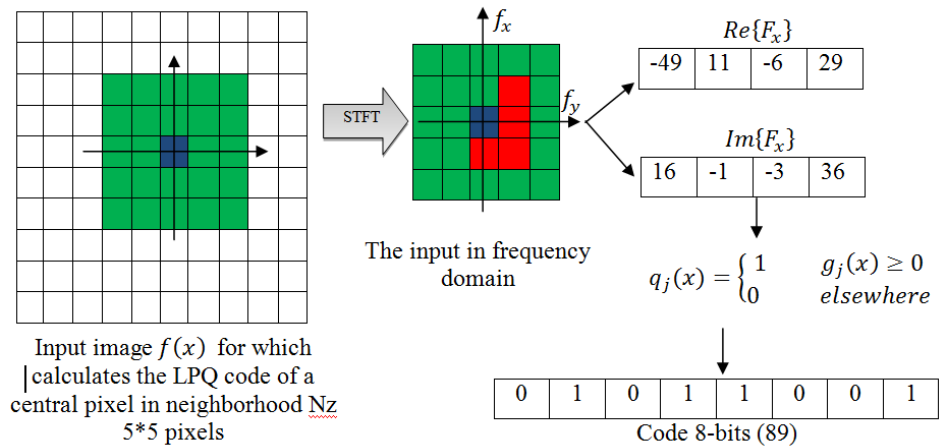

**Fig.2.** Local Phase Quantization

## Experimental results:

In this section two database, namely the QDF database and the USTC-NVIE database, are tested to check the proposed DNE method with the state of- art approaches: PCA, LPP, and LDE. All experiments employ the nearest-neighbor classifier. From each data base we have selected 35 random subjects and 3 images for each of them. Now, we have assumed different subspaces represents different quality of image. For example, subspace 1 represents good quality images, which are taken from 1m distance in QDF dataset, likewise subspace 2 represents little bit blur image taken at a distance of 3m. Each subject represents different class of object. The main objective is to increase the inter-class distance and at the same time decrease the intra-class distance, so that all the images (good quality and blur) belonging to a particular subject can come close to each other and hence classification will become easier.

## Conclusion:

Here the subspace learning method called discriminant neighborhood embedding (DNE) is applies for pattern classification. We suppose that multi-class data points in high-dimensional space tend to move due to local intra-class attraction or inter-class repulsion and the optimal embedding from the point of view of classification is discovered consequently. After being embedded into a low-dimensional subspace, data points in the same class form compact sub manifold whereas the gaps between submanifolds corresponding to different classes become wider than before. Experiments on the QDF and USTC-NVIE databases demonstrate the effectiveness of our method.